\documentclass[a4]{article}
\usepackage{graphicx}
\usepackage{hyperref}
\usepackage{amsmath, amssymb, amsfonts}
\usepackage{array}
\usepackage[lofdepth,lotdepth]{subfig}
\usepackage[ruled,vlined]{algorithm2e}

\begin{document}
\title{Decentralized learning with budgeted network load using Gaussian copulas and classifier ensembles}
%
%
\author{John~Klein\textsuperscript{1} \and
Mahmoud~Albardan\textsuperscript{1} \and
Benjamin~Guedj\textsuperscript{2} \and 
Olivier~Colot\textsuperscript{1}}
%
%
\date{\textsuperscript{1}Univ. Lille, CNRS, Centrale Lille, UMR 9189 - CRIStAL - Centre de Recherche en Informatique Signal et Automatique de Lille, F-59000 Lille \\
\{john.klein,mahmoud.albardan,olivier.colot\}@univ-lille.fr\\
\textsuperscript{2}Inria, France and University College London, United Kingdom\\
benjamin.guedj@inria.fr
}
\maketitle              
\begin{abstract}
We examine a network of learners which address the same classification task but must learn from different data sets. The learners cannot share data but instead share their models. Models are shared only one time so as to preserve the network load. We introduce DELCO (standing for Decentralized Ensemble Learning with COpulas), a new approach allowing to aggregate the predictions of the classifiers trained by each learner. The proposed method aggregates the base classifiers using a probabilistic model relying on Gaussian copulas. Experiments on logistic regressor ensembles demonstrate competing accuracy and increased robustness in case of dependent classifiers. A companion python implementation can be downloaded at \href{https://github.com/john-klein/DELCO}{https://github.com/john-klein/DELCO}.

\textbf{keywords}: Decentralized learning, classifier ensemble, copulas
\end{abstract}

Big data is both a challenge and an opportunity for supervised learning. 
It is an opportunity in the sense that we can train much more sophisticated models and automatize much more complex tasks. It is a challenge in the sense that conventional learning algorithms do not scale well when either the number of examples, the number of features or the number of class labels is large. On more practical grounds, it becomes also infeasible to train a model using a single machine for both memory and CPU issues. 

Decentralized learning is a setting in which a network of interconnected machines are meant to collaborate in order to learn a prediction function. Each node in the network has access to a limited number of training examples. Local training sets may or may not be disjoint and the cost of transferring all data to a single computation node is prohibitive. The cost of transfer should be understood in a general sense. It can encompass the network traffic load or the risk to violate data privacy terms. Decentralized learning is a framework which is well suited for companies or public institutions that wish to collaborate but do not want to share their data sets (partially of entirely).

There are several subfields in the decentralized learning paradigm that depend on the network topology and the granted data transfer budget. When any pairwise connection is allowed and when the budget is high, some well established algorithms can be adapted with limited effort to the decentralized setting. For instance, in deep neural networks \cite{dean2012large}, neural units can exchange gradient values to update their parameters as part of the backpropagation algorithm. This implies that some nodes are used just for training some given neural units or layers and do not have local training sets. The nodes that have training data must train the first layer and share their parameters. In the end, the amount of transferred data may in this case be greater than the entire training data transfer to a single node. 
When each node is meant to train a model from its private data set but nodes can only exchange symmetrically information with their one-hop neighbors in the network, Giannakis et al. \cite{giannakis2016decentralized} explain that the global optimization of the sum of losses over training data can be broken into several local optimization problems on each node. Since many training algorithms rely essentially on such an optimization problem, the method is rather generic. It also has the advantage that no training data has to be shared and that the distributed optimization can converge to the same parameter estimates as the global one. On the downside, the algorithm is iterative and the amount of transferred data cannot be anticipated. A similar decentralized learning problem is addressed in \cite{gholami2016decentralized} where an approximate Bayesian statistical solution is proposed.

In this article, we place ourselves in a context where the amount of transferred data must be anticipated and no training examples can be shared. We assume a fully connected topology allowing each node to share its trained base classifier with every other node as well as with a central node which will aggregate models. 
Local training phases do not have to be synchronized. Ensemble methods or multiple classifier systems are good candidates to operate in such a form of decentralized learning. Indeed, many such methods do not require that the base learners, \emph{i.e.} those trained on each local node, have to collaborate at training time.

In the central node, we train a probabilistic model to aggregate the base classifiers. We investigate a model relying on conditional probabilities of classifier outputs given the true class of an input (whose estimation can be decentralized without difficulty). These distributions are used as building blocks to classify unseen examples as those maximizing class probabilities given all classifiers outputs \cite{dawid1979maximum,kim2012bayesian}. The originality of our approach consists in resorting to copula functions to obtain a relatively simple model of joint conditional distributions of the local base classifier outputs given the true class. 

The next section presents the classifier aggregation problem and existing approaches addressing this issue. Section \ref{sec:method_outline} gives an outline of our new ensemble method. We first present this method in a centralized setting for simplicity. Its deployment in a decentralized setting is explained in the final subsection of this very section. Section \ref{sec:numerical_experiments} assesses the performances of this new ensemble method on both synthetic and real challenging data sets as compared to prior arts.


\subsection{Combining classifiers}
Let $\Omega$ denote a set of $\ell$ class labels $\Omega$ = \{$c_1,\hdots,c_\ell$\}, where each element $c_i$ represents one label (or class). 
Let $\mathbf{x}$ denote an input (or example) with $d$ entries. Most of the time, $\mathbf{x}$ is a vector and lives in $\mathbb{R}^d$ but sometimes some of its entries are categorical data and $\mathbf{x}$ lives in an abstract space $\mathbb{X}$ which does not necessarily have a vector space structure. For the sake of simplicity, 
we suppose in the following of this article that $\mathbf{x}$ is a vector.
    
A classification task consists in determining a prediction function $\hat{c}$ that maps any input $\mathbf{x}$ with its actual class $y$ $\in$ $\Omega$. 
This function is obtained from a training set $\mathcal{D}_{\textrm{train}}$ which contains pairs $\left(   \mathbf{x}^{(i)},y^{(i)}\right) $ where $y^{(i)}$ is the class label of example $\mathbf{x}^{(i)}$. The cardinality of the training set is denoted by  $n_{\text{train}}$. We usually also have a test set $\mathcal{D}_{\textrm{test}}$ which is disjoint from $\mathcal{D}_{\textrm{train}}$ to compute unbiased estimates of the prediction performance of function $\hat{c}$. The size of $\mathcal{D}_{\textrm{train}} \cup \mathcal{D}_{\textrm{test}}$ is denoted by $n$.

Given $m$ classifiers, the label $y$ assigned by the $k$\textsuperscript{th} classifier to the input $\mathbf{x}$ is denoted by $\hat{c}_k(\mathbf{x})$. 
In the usual supervised learning paradigm, each $\hat{c}_k$ is typically obtained by minimizing a weighted sum of losses incurred by deciding $\hat{c}_k \left( \mathbf{x}^{(i)} \right) $ as compared to $y^{(i)}$ for each data point in the training set or by building a function that predicts $y^{(i)}$ in the vicinity of $\mathbf{x}^{(i)}$ up to some regularity conditions.
Once we have trained multiple classifiers, a second algorithmic stage is necessary to derive an ensemble prediction function $\hat{c}_{\text{ens}}$ from the set of classifiers $\left\{ \hat{c}_1, \hdots, \hat{c}_m \right\}$.

Early attempts to combine classifiers focused on deterministic methods relying on voting systems \cite{Xu92} and Borda counts \cite{ho1994decision}. In later approaches \cite{Kit98,kittler2003sum}, some authors started to formalize the classifier aggregation problem in probabilistic terms when base classifier outputs are estimates of probabilities $p \left( y | \mathbf{x} \right) $. It is also possible to probabilistically combine classifiers without assuming that base classifiers rely themselves on probabilistic models. 
Indeed, we can picture the set of classifier predictions as entries of some vector $\mathbf{z}\left( \mathbf{x} \right)  = \left[ \hat{c}_1 \left( \mathbf{x} \right), \hdots ,\hat{c}_m \left( \mathbf{x} \right) \right]^T $. Regarding these vectors as new inputs, we resort to a decision-theoretic framework. 
Under 0-1 loss, the optimal decision rule (in terms of expected loss) is
     
\begin{equation}
    \hat{c}_{\text{ens}} \left(    \mathbf{x} \right)= \underset{y\in\Omega}{\arg\max}\hspace{0.2cm} p \left( y| \mathbf{z} \left( \mathbf{x} \right) \right) 
    .
    \label{eq:c_ens}
\end{equation}
Suppose we select $n_{\text{val}}$ training examples from $\mathcal{D}_{\text{train}}$ to build a validation set $\mathcal{D}_{\text{val}}$ and let $\mathcal{D}'_{\text{train}} = \mathcal{D}_{\text{train}} \setminus \mathcal{D}_{\text{val}}$. We can train functions $\hat{c}_1$ to $\hat{c}_m$ using $\mathcal{D}'_{\text{train}}$ and compute predictions for each member of the validation set. So we can build $n_{\text{val}}$ vectors $\mathbf{z}^{(i)}$ and use their labels $y^{(i)}$ to infer the parameters of the conditional distributions $p \left( y| \mathbf{z}  \right)$. 
In the next subsection, we detail such inference methods. 

Let alone probabilistic approaches, another possibility is to use the set of pairs $\left( \mathbf{z}^{(i)},y^{(i)} \right) $ to train a second stage classifier. This approach is known as stacking \cite{wolpert1992stacked} and has gained in popularity in the past decade as several machine learning competitions were won by stacked classifiers \cite{koren2009bellkor}. There are many other multiple classifier systems or ensemble methods in the literature but few of them are applicable in a decentralized setting. In particular, boosting \cite{freund1997decision} requires each ensemble component to see all data and bagging \cite{breiman1996bagging} consists in drawing bootstrap samples of training data so they would both require greater amounts of data transfer than simply sending all data to a single machine. To get a broader picture of the landscape of classifier combination and ensemble methods, we refer the reader to \cite{woz14}.

Stacking is also used in \cite{merz1998combining} along with correlation analysis in order to account for correlation in classifier predictions. Taking into account these correlations is the most important added value of the copula based probabilistic model that we introduce in section \ref{sec:method_outline}. The approach in \cite{merz1998combining} corresponds to a discriminative model while ours is a generative model of aggregation. It is not adapted to the decentralized setting as it involves a singular value decomposition of a matrix with $n\times m \times \ell$ entries which is prohibitive and propagates big data bottlenecks on the aggregation side.

\subsection{A probabilistic model of aggregation}

In this subsection, we present several approaches for inferring the parameters of the multinomial conditional distributions $p \left( y|\hat{c}_1 \left( \mathbf{x}) \right), \hdots ,\hat{c}_m \left( \mathbf{x} \right) \right)$. These approaches are essentially due to Dawid and Skene \cite{dawid1979maximum} and were promoted and further developed by Kim and Ghahramani \cite{kim2012bayesian} in the context of classifier combination; see also \cite{tresp2000bayesian} for a Bayesian committee algorithm tailored for Gaussian processes. 
Inferring parameters of multinomial distributions may not seem challenging at first sight. The problem is that, we need to solve $\ell^m$ such inference problems so the complexity of the problem does not scale well w.r.t. both $\ell$ and $m$. Applying Bayes formula, we have
    \begin{align}
    p\left(y|\hat{c}_1( \mathbf{x}),\hdots,\hat{c}_m( \mathbf{x})\right) \propto p\left(\hat{c}_1( \mathbf{x}),\hdots,\hat{c}_m( \mathbf{x}) | y\right) \times p \left( y \right). \label{eq:predictive}
    \end{align}
    The estimation of class probabilities is easy but again, the estimation of conditional joint distributions $p\left(\hat{c}_1( \mathbf{x}),\hdots,\hat{c}_m( \mathbf{x}) | y\right)$ has the same complexity as the estimation of the posterior. 
    
    Linear complexity can be achieved by making conditional independence assumptions that allow each conditional joint distribution to factorize as the product of its marginals, that is
    \begin{align}
    p\left(y|\hat{c}_1( \mathbf{x}),\hdots,\hat{c}_m( \mathbf{x})\right) \propto p \left( y \right) \times \prod _{i=1}^m p\left( \hat{c}_i( \mathbf{x})| y\right) . \label{eq:indep}
    \end{align}

In this approach, the parameters of $m+1$ multinomial distributions need to be estimated which does not raise any particular difficulty. Unfortunately, the independence assumption is obviously unrealistic: the classifier outputs are likely to be highly correlated. Indeed, examples that are difficult to classify correctly for classifier $\hat{c}_i$ are usually also difficult to classify correctly for any other classifier $\hat{c}_j$, $j\neq i$. 
The dependence between classifiers has its roots in several causes, such as learning on shared examples, use of classifiers of the same type, correlation between training examples. This accounts for the fact that misclassifications for each $\hat{c}_k$ occur most of the time with the same inputs. 
In spite of this, we will see that this approach achieves nice classification accuracy on several occasions. We believe this is explained by the same reason as the one behind naive Bayes classifier\footnote{This probabilistic approach can actually be regarded as a form of stacking in which the second stage classifier is a naive Bayes classifier.} efficiency. This model is an efficient technique although it also relies on unrealistic independence assumptions. Indeed, the inadequacy of these assumptions is compensated by a dramatic reduction of the number of parameters to learn making the technique less prone to overfitting.

Let us formalize the inference problem in a more statistical language to present further developments allowing to infer parameters in \eqref{eq:indep}. 
The classification output $\hat{c}_{k}$ of the $k$\textsuperscript{th} classifier is a random variable and the conditional distribution of $\hat{c}_{k}$ given $Y=y$ is multinomial: 
$\hat{c}_{k}|y \sim \textrm{Mult}\left( \boldsymbol\theta^{(k)}_y \right)$ with $\boldsymbol\theta^{(k)}_y$ a parameter vector of size $\ell$: $\boldsymbol\theta^{(k)}_y  = \left[ \theta^{(k)}_{y,1} \hdots \theta^{(k)}_{y,\ell} \right]^T $. 
In other words, the success/failure probabilities of the $k$\textsuperscript{th} classifier are the parameters $\theta_{y,i}^{(k)} = p \left( \hat{c}_k = i |y \right) $. The random variable $Y$ representing class labels has a multinomial distribution as well: $Y \sim \textrm{Mult}\left( \boldsymbol\gamma \right)$ and $\boldsymbol\gamma$ is another vector of parameters of size $\ell$. 
Let $\mathcal{D}_{\textrm{agg}}$ denote the data set whose elements are tuples $\left( \hat{c}_{1}\left(\mathbf{x}^{(i)}\right),\hdots,\hat{c}_{m}\left( \mathbf{x}^{(i)}\right),y^{(i)}\right)$ for $\left( \mathbf{x}^{(i)}, y^{(i)}\right) \in \mathcal{D}_{\textrm{val}}$.
Under classifier independence assumptions, the likelihood writes  
    \begin{equation}
     p \left(\mathcal{D}_{\textrm{agg}} | \boldsymbol\theta^{(1)}_1,\hdots,\boldsymbol\theta^{(m)}_\ell,\boldsymbol\pi \right) = \prod_{i=1}^{n_{\textrm{val}}} \gamma_{y^{(i)}} \prod_{k=1}^m \theta^{(k)}_{y^{(i)},i_k},
    \end{equation}
where $i_k = \hat{c}_k \left( \mathbf{x}^{(i)} \right) $. 
Maximum likelihood estimates of $\boldsymbol\gamma$ and each $\boldsymbol\theta^{(k)}_{y}$ are known in closed form and can be easily computed. Kim and Ghahramani \cite{kim2012bayesian} propose a Bayesian treatment consisting of using hierarchical conjugate priors on the parameters of all conditional distributions $p\left( c_{k}|y \right)$ as well as on the class distribution $p\left( y\right)$. The conjugate priors for $\boldsymbol\theta^{(k)}_y$ and $\boldsymbol\gamma$ are Dirichlet: 
    $\boldsymbol\theta^{(k)}_y \sim \textrm{Dir}\left( \boldsymbol\alpha^{(k)}_y\right)$ and $\boldsymbol\gamma \sim \textrm{Dir}\left( \boldsymbol\beta \right)$.
A second level of priors is proposed for the parameters $\boldsymbol\alpha^{(k)}_y$. The conjugate prior distribution of each $\boldsymbol\alpha^{(k)}_y$ is exponential.
Gibbs and rejection sampling are then used to infer these parameters.
    
Finally, Kim and Ghahramani \cite{kim2012bayesian} also extend this model in order to take into account dependencies between classifiers. They propose to use a Markov random field as a model of classifier output interactions. The main limitation of this method is the high computational cost induced by MCMC and rejection sampling. In the next section, we introduce a copula-based model that allows to grasp classifier dependency without resorting to an MCMC step.

\section{Method outline} 
\label{sec:method_outline}

In this section, we present a new ensemble method allowing to build the decision function $\hat{c}_{\text{ens}}$ from \eqref{eq:predictive} without resorting to some conditional independence assumption. 
We propose a Gaussian copula model for the conditional joint distributions $p\left(\hat{c}_1( \mathbf{x}),\hdots,\hat{c}_m( \mathbf{x}) | y\right)$. We start by giving elementary background on copulas and later explain how they can be efficiently implemented in a decentralized learning setting. 

\subsection{Copulas} 
\label{sub:copulas}
An $m$-dimensional copula function $\text{Cop}: \left[ 0;1 \right]^m \rightarrow \left[ 0;1 \right]  $ is a cumulative distribution with uniform marginals. The growing popularity of these functions stems from Sklar's theorem which asserts that, for every random vector $\mathbf{L} \sim f$, there exist a copula $\text{Cop}$ such that $F = \text{Cop} \circ \mathbf{G}$ where $F$ is the cumulative version of distribution $f$ and $\mathbf{G}$ is a vector whose entries are the cumulative marginals $G_k \left( a \right)  = F \left( \infty,\hdots, \infty, a ,\infty, \hdots, \infty \right) $ for any $a$ in the $k$-dimensional domain of $f$. 

When $F$ is continuous, the copula is unique. When we deal with discrete random variables as in our classification problem, the non-uniqueness of the copula raises some identifiability issues \cite{genest2007primer,faugeras2017inference}. Without denying the importance of these issues, we argue that, from a pattern recognition standpoint, what essentially matters is to learn a model that generalizes well. For instance, there are also identifiability issues for neural networks \cite{sontag1998learning} which do not prevent deep nets to achieve state-of-the-art performance in many applications.

In this article, we investigate parametric copula families to derive a model for the conditional joint distributions $p\left(\hat{c}_1( \textbf{X}),\hdots,\hat{c}_m( \textbf{X}) | y\right)$ where $\textbf{X}$ is the random vector capturing input uncertainty. Parametric copulas with parameters vector $\lambda$ are denoted by $\text{Cop}_{\lambda}$. A difficulty in the quest for an efficient ensemble method is that we must avoid working with cumulative distributions because the computational cost to navigate from cumulative to non-cumulative distributions is prohibitive. We can compute Radon-Nikodym derivatives of $\text{Cop}_{\lambda} \circ \mathbf{G}$ w.r.t. a reference measure but again since we work in a discrete setting we will not retrieve closed form expression for $f$ for an arbitrary large number of classifiers. 
As a workaround, we propose to embed each discrete variable $\hat{c}_k \left( \textbf{X} \right)|y $ in the real interval $\left[ 0;\ell \right[ $. 
Let $f_y: \mathbb{R}^m \rightarrow \mathbb{R}^+$ be a probability density (w.r.t. Lebesgue) whose support is $\left[ 0;\ell \right[^m$ and such that for any $\mathbf{z} \in \Omega^m$, we have $f_y \left( \mathbf{a} \right) = p \left( \hat{c}_1 \left( \textbf{X}\right)=z_1, \hdots, \hat{c}_m \left( \textbf{X} \right)=z_m |y \right) $ for any vector $\mathbf{a}$ in the unit volume $\mathcal{V}_{\mathbf{z}} = \left[ z_1 - 1;z_1 \right[ \times \hdots \times \left[ z_m- 1;z_m \right[$. 
This means that $f_y$ is piecewise constant and it can be understood as the density of some continuous random vector whose quantized version is equal in distribution to the tuple $\left( \hat{c}_1 \left( \textbf{X}\right)|y, \hdots, \hat{c}_m \left( \textbf{X}\right) |y\right) $. 
Moreover, if $f_y^{(i)}$ is the $i$\textsuperscript{th} marginal density of $f_y$, we also have $f_y^{(i)} \left( a \right) =  p \left( \hat{c}_1 \left( \textbf{X}\right)=z|y \right)$ for any $a \in \left[ z-1;z \right[$ and any $z \in \left\{ 1; \hdots; \ell \right\}$. 
For any $\mathbf{z} \in \Omega^m$, according to this continuous random vector vision of the problem, we can now thus write 
\begin{align}
p\left(\hat{c}_1=z_1,\hdots,\hat{c}_m=z_m | y\right) &= \text{cop}_{\lambda} \left(  \mathbf{u} \right) \times \prod_{i=1}^m  p \left( \hat{c}_i=z_i|y \right)  \label{eq:joint_cop}, \\ 
\mathbf{u} &= \left[ F_{1,y} \left( z_1 \right) ,\hdots, F_{m,y} \left( z_m \right)  \right]
\end{align}
where $\text{cop}_{\lambda}$ is the density of $\text{Cop}_{\lambda}$ and $F_{i,y}$ is the cumulative distribution of variable $\hat{c}_i \left( \textbf{X} \right)|y $. 
This construction is not dependent in the (arbitrary) way in which the elements of $\Omega$ are indexed. 

Among parametric copula families, the only one with a closed form density for arbitrary large $m$ is the Gaussian copula. The density of a Gaussian copula \cite{ZEZULA2009} is given by 
\begin{align}
\text{cop}_{\lambda} \left( \mathbf{u} \right) &= \frac{1}{|\mathbf{R}|^{1/2}} \exp \left( - \frac{1}{2} \mathbf{v}^T \cdot \left(  \mathbf{R}^{-1} - \mathbf{I}  \right)\cdot \mathbf{v} \right) ,
\end{align}
where $\mathbf{R}$ is a correlation matrix, $\mathbf{I}$ is the identity matrix and $\mathbf{v}$ is a vector with $m$ entries such that $v_k = Q \left( u_k \right) $ where $Q$ is the quantile function of a standard normal distribution. The copula parameter in this case is the correlation matrix. Estimating the entries of this matrix is not trivial. We will therefore choose a simplified model and take $\mathbf{R} = \lambda \mathbf{1} +  \left( 1 - \lambda \right)\mathbf{I}$ where $\mathbf{1}$ is the all-one matrix. In this model, each diagonal entry of $\mathbf{R}$ is $1$ and each non-diagonal entry is $\lambda$. The dependency between classifier outputs is regulated by $\lambda$ which is a scalar living in $\left( \frac{-1}{m-1} ;1 \right)$. We also make the assumption that correlation matrices are tied across conditionings on $Y=y$. 
The $m \times \ell$ cumulative distributions $F_{i,y}$ are evaluated using estimates of the vectors $\boldsymbol \theta^{(i)}_y = \left[ p \left( \hat{c}_i=c_1|y \right) \hdots p \left( \hat{c}_i=c_\ell|y \right) \right]^T $.

Observe that when $\lambda=0$, the copula density is constant one and the proposed model boils down to the independent case \eqref{eq:indep}. Since our model is a generalization of \eqref{eq:indep}, this latter is referred to as the independent copula-based ensemble in the remainder of this article but it should be kept in mind that it is a prior art.


\subsection{New ensemble method} 
\label{sub:new_ensemble_method}
Now that we have introduced all the ingredients to build our new ensemble method, let us explain how it can be implemented efficiently in practice. The only crucial remaining problem is to tune the parameter $\lambda$ of the parametric copula. This parameter summarizes the dependency information between each pair of random variables $\left( \hat{c}_k \left( \textbf{X} \right)|y\, ;\, \hat{c}_{k'} \left( \textbf{X} \right)|y  \right) $. 

Since we have only one parameter to set, we can use a grid search on the interval $\left( \frac{-1}{m-1} ;1 \right)$ using the validation set and select $\hat{\lambda}$ as the value achieving maximal accuracy on this validation set. In the experiments, we use an evenly spaced grid (denoted $\text{grid}_\lambda$) containing 101 values. 
In the sequel, our approach will be referred to as Decentralized Ensemble Learning with COpula (DELCO). The pseudo-code for DELCO is given in Algorithm \ref{al:train_algo}.

    \begin{algorithm}[t]
    \DontPrintSemicolon
     \caption{DELCO (training)}\label{al:train_algo}
     \KwData{$\mathcal{D}_\text{train}$, $n_\text{val}$, $\text{grid}_\lambda$ and $\left\{ \text{train-alg}_k \right\}_{k=1}^m$}
     Select $n_\text{val}$ data points from $\mathcal{D}_\text{train}$ to build $\mathcal{D}_\text{val}$ \;
     $\mathcal{D}'_\text{train} \leftarrow \mathcal{D}_\text{train} \setminus \mathcal{D}_\text{val}$\;
     \For{$k\in \left\{ 1,\hdots, m \right\}$}{
      Run $\text{train-alg}_k$ on $\mathcal{D}'_\text{train}$ to learn $\hat{c}_k$ \;
     }
     
     \For{$y\in \left\{ 1,\hdots, \ell \right\}$}{
        $\gamma_y \leftarrow \frac{1+\sum\limits_{i=1}^{n_{\textrm{val}}} \mathbb{I}_{y}\left(y^{(i)}\right)}{\ell+ n_{\textrm{val}}}$\;
        \For{$k\in \left\{ 1,\hdots, m \right\}$}{
          \For{$j\in \left\{ 1,\hdots, \ell \right\}$}{
            $\theta^{(k)}_{y,j} \leftarrow \frac{1+\sum\limits_{i=1}^{n_{\textrm{val}}} \mathbb{I}_{y}\left(y^{(i)}\right) \mathbb{I}_{j}\left(\hat{c}_k\left(\mathbf{x}^{(i)}\right)\right) }{\ell + \sum\limits_{i=1}^{n_{\textrm{val}}} \mathbb{I}_{y}\left(y^{(i)}\right)}$\;
            $F_{k,y} \left( j \right) \leftarrow \left[ 1 -  \mathbb{I}_0 \left( j \right) \right] \times  F_{k,y} \left( j-1 \right) + \theta^{(k)}_{y,j} $\;
          }
        
        }
     }
     
     \For{$\lambda \in \text{grid}_\lambda$}{
        Obtain $\hat{c}_{\text{ens}}$ by substituting \eqref{eq:joint_cop} in \eqref{eq:predictive} and then \eqref{eq:predictive} in \eqref{eq:c_ens}, and  using $\hat{c}_1,\hdots, \hat{c}_m, \boldsymbol\gamma, \boldsymbol\theta^{(1)}_1, \hdots, \boldsymbol\theta^{(m)}_\ell$ and $\lambda$\;
        $\text{Acc} \left( \lambda \right) \leftarrow \frac{\sum\limits_{i=1}^{n_{\textrm{val}}} \mathbb{I}_{y^{(i)}}\left( \hat{c}_{\text{ens}} \left(  \mathbf{x}^{(i)} \right)  \right) }{n_{\textrm{val}}}$ \;
     }
     $\hat{\lambda} \leftarrow \underset{\lambda \in \text{grid}_\lambda}{\arg\max}\; \text{Acc} \left( \lambda \right)$\;
     Obtain $\hat{c}_{\text{ens}}$ by substituting \eqref{eq:joint_cop} in \eqref{eq:predictive} and then \eqref{eq:predictive} in \eqref{eq:c_ens}, and using $\hat{c}_1,\hdots, \hat{c}_m, \boldsymbol\gamma, \boldsymbol\theta^{(1)}_1, \hdots, \boldsymbol\theta^{(m)}_\ell$ and $\hat{\lambda}$\;
     \Return{$\hat{c}_{\text{ens}}$}
     
     \label{algo}
     \end{algorithm}
     

In Algorithm \ref{al:train_algo}, $\mathbb{I}_x$ denotes the indicator function of the singleton $\{x\}$. 
The vectors of parameters $\boldsymbol\gamma$ and $\left\{ \boldsymbol\theta^{(1)}_1, \hdots, \boldsymbol\theta^{(m)}_\ell \right\}$ are estimated using the Laplace add-one smoothing which is the conditional expectation of the parameters given the data in a Dirichlet-multinomial model. As opposed to maximum likelihood estimates, it avoids zero counts which are numerically speaking problematic. It is also recommended to maximize the log-version of \eqref{eq:c_ens} which is numerically more stable.

Finally, one can optionally retrain the classifiers on $\mathcal{D}_\text{train}$ after $\hat{\lambda}$ is estimated. Since $\mathcal{D}_\text{train}$ is larger then $\mathcal{D}'_\text{train}$, it allows training algorithms to converge to possibly slightly better decision functions. Training them initially on $\mathcal{D}_\text{train}$ is however ill-advised as the parameter estimates would be biased. In the next section, where we present numerical results, we use this optional step.

\subsection{Dencentralized DELCO} 
\label{sub:dencentralized_delco}
In the previous paragraphs, we presented our new ensemble method in the centralized setting first for simplicity. It can be adapted to the decentralized setting described in the introduction with little efforts. To achieve decentralized learning with DELCO, each local private data set needs to be separated in a local training set and a local validation set. After all locally trained models are exchanged between all nodes, each node computes the confusion matrix of each base classifier using its local validation set. These matrices are sent to the central node which just needs to average and normalize them to obtain the estimates of vectors $\boldsymbol \theta^{(i)}_y$. Similarly, vector $\boldsymbol\gamma$ can be estimated by sending to the central node the number of examples belonging to each class. 
Finally, grid search can also be implemented in the same fashion. The central node can send the global estimates of $\boldsymbol \theta^{(i)}_y$ and $\boldsymbol\gamma$ to each node. Each node can then perform grid search using its local validation set, compute accuracies and send them back to the central node which will average them. Note that the number and the cost of transfers through the network are known before starting to train. 




\section{Numerical experiments} 
\label{sec:numerical_experiments}

In this section, the performance of DELCO is assessed in terms of classification accuracy and robustness. 
Situations in which aggregation performance discrepancies are most visible usually occur when there is diversity \cite{hansen1990neural,krogh1995neural} in the trained base prediction functions $\hat{c}_i$. Among other possibilities \cite{ali1995link,merz1995dynamic,opitz1996generating,maclin1995combining}, one way to induce diversity consists in distributing data points across the network of base classifiers in a non-iid way, that is, each base classifiers only sees inputs that belong to a given region of the feature space. This is a realistic situation as the data stored in a network node might be dependent on the geographic location of this node for instance.

Furthermore, we chose to combine base classifiers with limited capacity, \emph{i.e.} \emph{weak} classifiers as in boosting \cite{freund1997decision}, so that the aggregated model has a significantly larger capacity allowing to discover better decision frontiers. We decided to use logistic regression on each local data set as this algorithm yields a linear decision frontier. Also, logistic regression has the advantage to have no hyperparameter to tune making the conclusions from the experiments immune to this issue. This is also the reason why we do not use a regularized version of this algorithm.

In each experiment, $10\%$ of the data are used for validation, i.e. $n_{\text{val}} = \frac{n_{\text{train}}}{10}$. We compare DELCO to the following state-of-the-art or reference methods:
\begin{itemize}
  \item classifier selection based on accuracies,
  \item best base classifier,
  \item weighted vote combination based on accuracies,
  \item stacking,
  \item centralized classifier trained on all data,
  \item the independent copula ensemble (equivalent to \eqref{eq:indep}). 
\end{itemize}

Each method relying on base classifier accuracies uses estimates obtained from the validation set. The validation set is also used as part of stacking to generate inputs for the second stage training. We also use a logistic regression for this second stage and input entries are predicted classes from each base classifier. Stacking is applicable if the validation set is shared across learners. The best base classifier and the centralized classifier are relevant references to assess the quality of the aggregation. 
Concerning DELCO, we examine the simplified Gaussian copula where the copula hyperparameter is estimated by grid search from the validation set. In a reproducible research spirit, we provide a python implementation of DELCO and other benchmarked methods (\href{https://github.com/john-klein/DELCO}{https://github.com/john-klein/DELCO}). 

\subsection{Synthetic data} 
\label{sub:synthetic_data}
\begin{figure}[!t]
\centering
\subfloat[Moons]{\includegraphics[width=.33\textwidth]{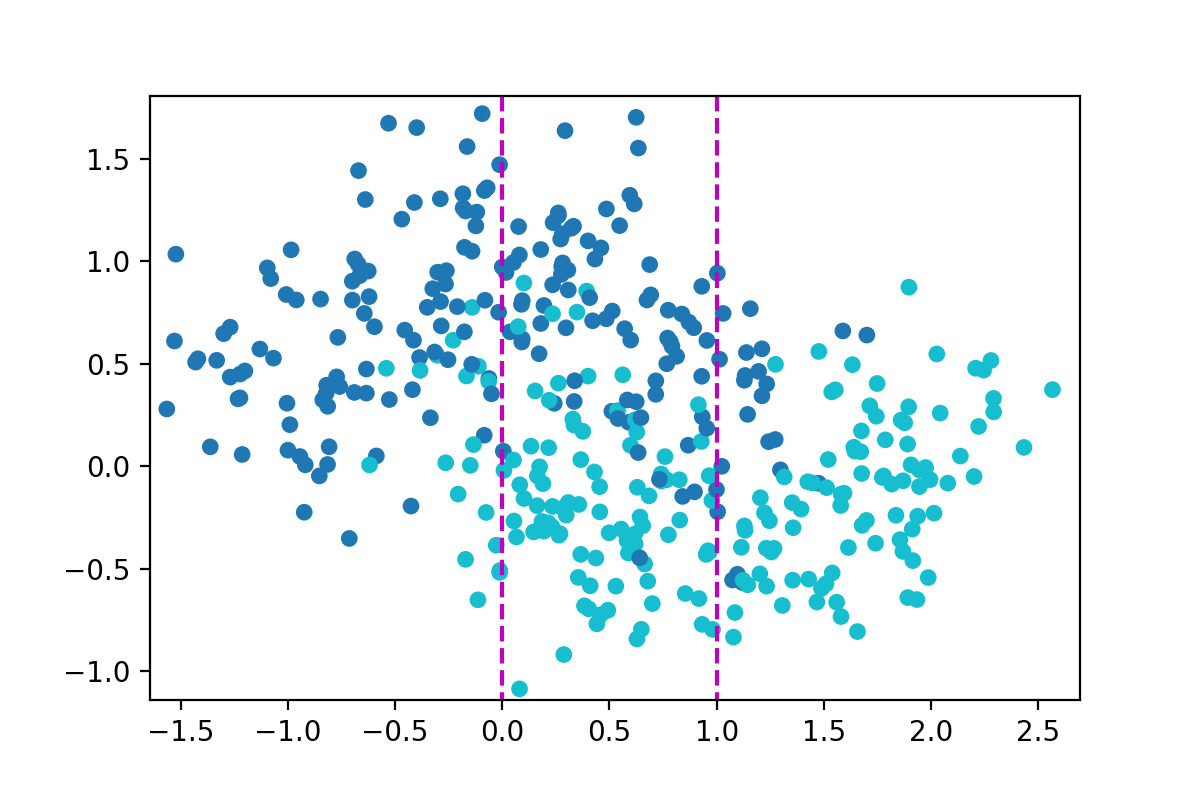}\label{subfig:moons}}
\subfloat[Blobs]{\includegraphics[width=.33\textwidth]{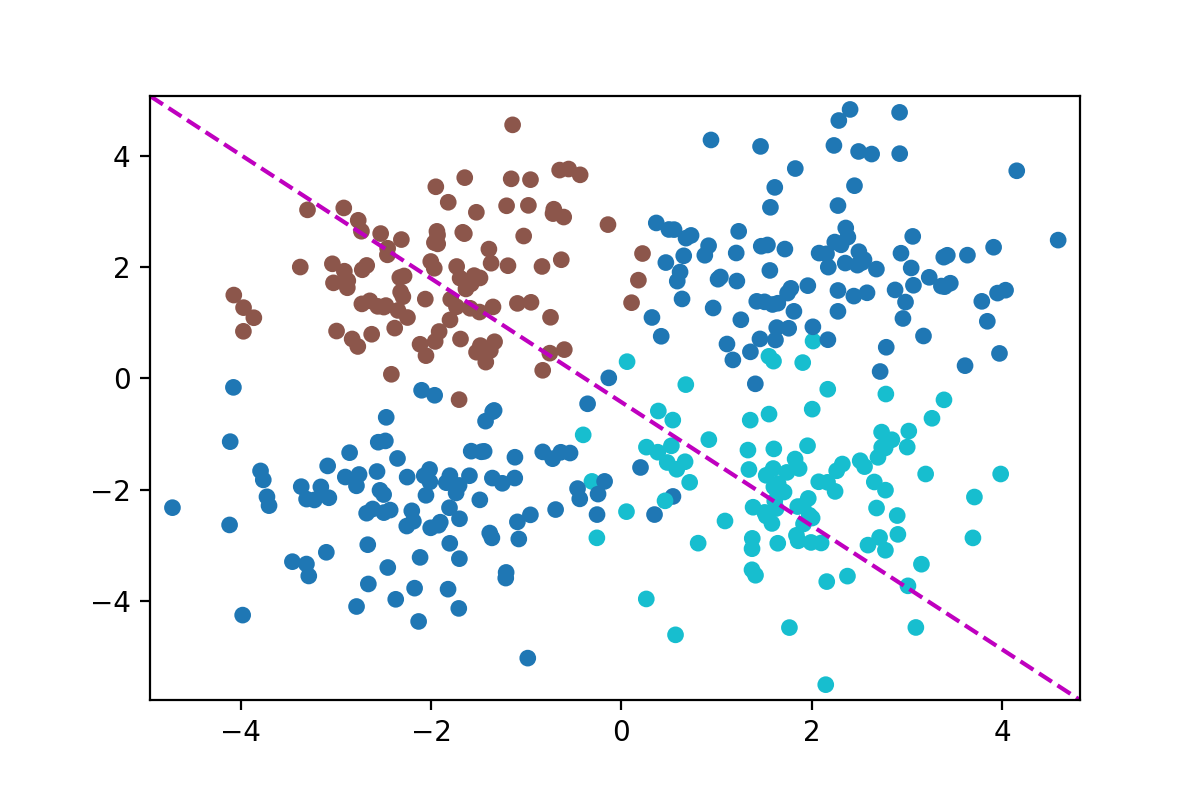}\label{subfig:blobs}} 
\subfloat[Circles]{\includegraphics[width=.33\textwidth]{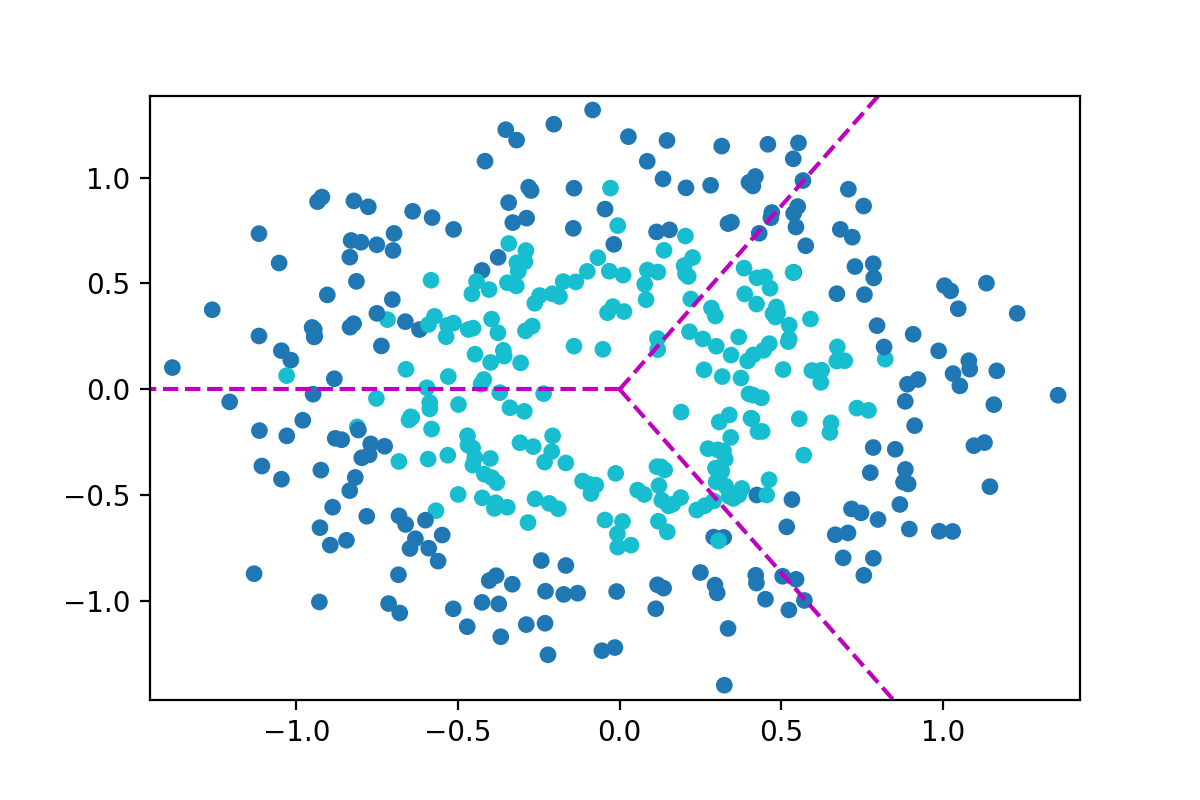}\label{subfig:circles}}
\caption{Synthetic data sets and their partitions into feature space regions ($n=400$).}
\label{fig:synth_datasets}
\end{figure}

Using synthetic data sets is advantageous in the sense that, in the test phase, we can generate as many data as we want to obtain very reliable estimates of classification accuracies. We examine three different data generation processes from \href{the http://scikit-learn.org/stable/}{\texttt{sklearn}} library \cite{scikit-learn}: Moons, Blobs and Circles. Each of these processes yields non-linearly separable data sets as illustrated in Figure \ref{fig:synth_datasets}. 

The Moons and Circles data sets are binary classification problems while Blobs involves three classes. For each problem, the data set is partitioned into disjoint regions of the input space as specified in Figure \ref{fig:synth_datasets} and consequently we combine two base classifiers for the Blobs data set and three base classifiers for the others. Also, in each case, input vectors live in $\mathbb{R}^2$.

The Moons data set consists in two half-circles to which a Gaussian noise is added. For each half-circle, one of its extremal point is the center of the other half-circle. The covariance matrix of the noise in our experiment is $0.3 \times \mathbf{I}$ where $\mathbf{I}$ is the identity matrix. Before adding this noise, we also randomized the position of sample points on the half circle using a uniform distribution while the baseline sklearn function samples such points with fixed angle step.
The Blobs data set is also obtained using a slightly different function than its sklearn version. It generates a data set from four 2D Gaussian distributions centered on each corner of a centered square whose edge length is 4. Each distribution covariance matrix is $\mathbf{I}$. The examples generated by the distributions whose expectations are $\left( -2;-2 \right) $ and $\left( 2;2 \right) $ are assigned to class $c_0$. Each remaining Gaussian distribution yields examples for either class $c_1$ or $c_2$. 
Finally, the Circles data set consists in sampling with fixed angle step two series of points from centered circles with radius $0.5$ and $1$. A Gaussian noise with covariance matrix $0.15\times \mathbf{I}$ is added to these points. The python code for the synthetic data set generation is also online.

To evaluate the accuracy of a classifier or classifier ensemble trained on a data set drawn from any of the above mentioned generating processes, we drew test points from the same process until the Clopper-Pearson confidence interval of the accuracy has length below $0.2 \%$ with confidence probability $0.95$. For each generating process, we repeated this procedure $3000$ times to estimate the expected accuracy across data set draws.

\begin{table}[!t]

\renewcommand{\arraystretch}{1.2}
\caption{Classification accuracies for several synthetic data sets. ($n_{\text{train}}=200$ in the left table, $n_{\text{train}}=400$ in the right table)} 
\label{tab:synth1}
\centering
\resizebox{.48\textwidth}{!}{
\begin{tabular}{|c||c|c|c|}
\hline
 Method & Moons & Blobs & Circles\\
\hline
Clf. Selection & $79.25\%$ & $72.34\%$ & $62.38\%$ \\
& std. $3.51\%$ & std. $0.37\%$ & std. $0.32\%$ \\
\hline
Best Clf.  & $79.25\%$ & $72.34\%$ & $62.38\%$ \\
& std. $1.67\%$ & std. $0.36\%$ & std. $0.32\%$\\
\hline
Weighted Vote & $84.60\%$ & $82.43\%$ & $50.50\%$ \\
& std. $2.20\%$ & std. $11.02\%$ & std. $0.05\%$ \\
\hline
Stacking & $81.07\%$ & $69.87\%$ & $70.20\%$ \\
& std. $3.89\%$ & std. $5.37\%$ & std. $8.08\%$\\
\hline
Indep. Copula & $83.46\%$ & $91.14\%$ & $79.32\%$ \\
 & std. $2.91\%$ & std. $7.27\%$ & std. $6.70\%$ \\
\hline
DELCO & $80.57\%$ & $93.15\%$ & $84.49\%$ \\
Gauss. Copula & std. $4.68\%$ & std. $4.83\%$ & std. $4.51\%$ \\
\hline
Centralized Clf. & $84.99\%$ & $88.49\%$ & $50.02\%$ \\
& std. $0.55\%$ & std. $0.42\%$ & std. $0.49\%$\\
\hline
Optimal & $91.50\%$ & $95.50\%$ & $94.50\%$ \\
& std. $0\%$ & std. $0\%$ & std. $0\%$ \\
\hline
\end{tabular}} 
\resizebox{.48\textwidth}{!}{
\begin{tabular}{|c||c|c|c|}
\hline
 Method & Moons & Blobs & Circles\\
\hline
Clf. Selection & $79.67\%$ & $72.43\%$ & $62.50\%$ \\
& std. $2.14\%$ & std. $0.27\%$ & std. $0.05\%$ \\
\hline
Best Clf.  & $80.66\%$ & $72.45\%$ & $62.50\%$ \\
& std. $1.08\%$ & std. $0.22\%$ & std. $0.06\%$\\
\hline
Weighted Vote & $87.83\%$ & $78.72\%$ & $50.50\%$ \\
& std. $1.19\%$ & std. $9.96\%$ & std. $0\%$ \\
\hline
Stacking & $85.32\%$ & $71.70\%$ & $78.19\%$ \\
& std. $4.08\%$ & std. $2.61\%$ & std. $6.95\%$\\
\hline
Indep. Copula & $86.43\%$ & $93.78\%$ & $84.54\%$ \\
 & std. $3.28\%$ & std. $2.48\%$ & std. $4.45\%$ \\
\hline
DELCO & $86.75\%$ & $94.39\%$ & $86.39\%$ \\
Gauss. Copula & std. $3.07\%$ & std. $0.96\%$ & std. $1.11\%$ \\
\hline
Centralized Clf. & $85.22\%$ & $88.72\%$ & $50.01\%$ \\
& std. $0.45\%$ & std. $0.42\%$ & std. $0.50\%$\\
\hline
Optimal & $91.50\%$ & $95.50\%$ & $94.50\%$ \\
& std. $0\%$ & std. $0\%$ & std. $0\%$ \\
\hline
\end{tabular}}

\end{table}


The estimated expected accuracies and the estimated accuracy standard deviations are given for each classification method of the benchmark in Table \ref{tab:synth1} for $n_{\text{train}}=200$ and $n_{\text{train}}=400$. In these experiments, one of the copula-based methods is the top 2 method for the Moons data set and is the top 1 for the Blobs and Circles data sets. Most importantly, both copulas based method are obviously more robust since they never perform poorly on any data set. While the weighted vote method is the top 1 for Moons data set, it completely crashed on the Circles data set and converges to a random classifier.

Another result which is surprising at first sight, is that the centralized classifier is sometimes outperformed by some decentralized ensembles. This is actually well explained by the deterministic way in which input spaces are partitioned. Indeed, the partitions are cleverly chosen so that a combination of linear decision frontiers fits intuitively a lot better the data than a single linear separation does. In other words, ensembles have a larger VC dimension and visit a larger hypotheses set. One may wonder to which extent it would be possible to purposely partition data sets in such a relevant way to reproduce such conditions in more general situations. This is however beyond the scope of this article in which we address decentralized learning, a setting where we take distributed data as is and we cannot reorganize them.


There are three situations in which significant performance discrepancies are observed between DELCO and the independent copula. The first one is the Moons data set when $n_{\text{train}}=200$. We argue that DELCO fails to correctly estimate the parameter $\lambda$ as performance levels are reversed when $n_{\text{train}}=400$ and the validation set has now 40 elements instead of 20. 

The other situations are the Circles data set when either $n_{\text{train}}=200$ or $n_{\text{train}}=400$. In this case, we see that the independent copula-based ensemble fails to keep up with DELCO regardless of how many points the validation set contains. In conclusion, DELCO does offer increased robustness as compared to the independent copula model provided that the validation set size allows to tune correctly $\lambda$. Remember that when $\lambda=0$, both models coincide, so if we have enough data and if being independent is really what works best, then there is no reason why we should not obtain $\hat{\lambda}=0$. 


\subsection{Real data} 
\label{sub:real_data}

To upraise the ability of the benchmarked methods to be deployed in a decentralized learning setting, we also need to test them on sets of real data. Since decentralized learning is essentially useful in a big data context, we chose eight from moderate to large public data sets. The specifications of these data sets are reported in Table \ref{tab:real_specs}. 

\begin{table}[!t]
\renewcommand{\arraystretch}{1.2}
\caption{Real data set specifications}
\label{tab:real_specs}
\centering
\resizebox{.99\textwidth}{!}{
\begin{tabular}{|>{\centering\let\newline\\\arraybackslash}p{.20\textwidth}||>{\centering\let\newline\\\arraybackslash}p{.12\textwidth}|>{\centering\let\newline\\\arraybackslash}p{.20\textwidth}|>{\centering\let\newline\\\arraybackslash}p{.15\textwidth}|>{\centering\let\newline\\\arraybackslash}p{.14\textwidth}|>{\centering\let\newline\\\arraybackslash}p{.3\textwidth}|}
\hline
 Name   & Size $n$ & Dim. $d$ & Nbr. of classes $\ell$ & Data type &  Source \\
 \hline
 20newsgroup   & $18846$ & $100$ (after red.) & 20 & text & sklearn \\
 \hline
 MNIST & $70000$ & $784$ & $10$ & image & sklearn  \\
 \hline
 Satellite & $6435$ & $36$ & $6$ & image features & UCI repo. (Statlog) \\
 \hline
 Wine & $6497$ & $11$ & $2$ (binarized) & chemical features & UCI repo. (Wine Quality) \\
 \hline
 Spam & $4601$ & $57$ & $2$ & text & UCI repo. (Spam) \\
 \hline
 Avila & $10430$ & $10$ & $2$ (binarized) & layout features & UCI repo. (Avila) \\
 \hline
 Drive & $58509$ & $48$ & $11$ & current statistics & UCI repo. (Sensorless Drive Diagnosis) \\
 \hline
 Particle & $130064$ & $50$ & $2$ & signal & UCI repo. (MiniBooNE particle identification) \\
 \hline
\end{tabular}

}
\end{table}

Example entries from the 20newsgroup data set are word counts obtained using the term frequency - inverse document frequency statistics. We reduced the dimensionality of inputs using a latent semantic analysis \cite{deerwester1990indexing} which is a standard practice for text data. We kept 100 dimensions. Also, as recommended, we stripped out each text from headers, footers and quotes which lead to overfitting. 
Besides, for the Wine and Avila datasets, the number of class labels is originally 10 and 12 respectively. We binarized these classification tasks because some classes have very small cardinalities making it impossible for each node to have access to at least one example of this class. Aggregating base classifiers trained w.r.t different subsets of class labels goes behind the scope of this paper and will be touched in future works. In the Wine data set, class labels are wine quality scores. Two classes are obtained by comparing scores to a threshold of 5. In the avila dataset, class labels are middle age bible copyist identities. The five first copyists are grouped in one class and the remaining ones in the other class.

Unlike synthetic data sets, we need to separate the original data set into a train set and a test set. To avoid a dependency of the reported performances w.r.t train/test splits, we perform 2-fold cross validation (CV). Also, we shuffled at random examples and repeated the training and test phases $100$ times.

To comply with the diversity condition, we distributed the training data over network nodes using the following procedure: for each data set, for each class, 
\begin{enumerate}
   \item apply principal component analysis to the corresponding data,
   \item project this data on the dimension with highest eigenvalue,
   \item sort the projected values and split them into $m$ subsets of cardinality $n_i/m$ where $n_i$ is the proportion of examples belonging to class $c_i$.
 \end{enumerate} 
Each such subset is sent to only one node (the node being chosen arbitrarily). We argue that this way of splitting data is somehow adversarial because some nodes may see data that are a lot easier to separate than it should and will consequently not generalize very well. Average accuracies over random shuffles and CV-folds are given in Table \ref{tab:real2} for $m =10$ nodes.


\begin{table}[!t]
\renewcommand{\arraystretch}{1.2}
\caption{Classification accuracies (with standard deviations) for several real data sets. ($m=10$ nodes) }
\label{tab:real2}
\centering
\resizebox{\textwidth}{!}{
\begin{tabular}{|c||c|c|c|c|c|c|c|c|}
\hline
 Method & 20newsgroup & MNIST & Satellite & Wine & Spam & Avila & Drive & Particle \\
\hline
Clf. Selection & $37.35\%$ & $66.26\%$ & $77.83\%$ & $63.23\%$ & $85.26\%$ & $60.76\%$  & $58.58\%$  & $81.28\%$ \\
& std. $1.38\%$ & std.  $1.57\%$ & std. $2.04\%$ & std. $5.51\%$ & std. $1.31\%$ & std. $3.80\%$  & std. $2.77\%$ & std. $1.07\%$ \\
\hline
Best Clf.  & $38.25\%$ & $67.24\%$ & $79.10\%$ & $64.83\%$ & $86.60\%$  & $62.79\%$  & $58.77\%$  & $81.81\%$ \\
& std. $0.68\%$ & std. $0.76\%$ & std. $1.16\%$ & std. $4.75\%$ & std. $1.32\%$ & std. $2.24\%$  & std. $2.60\%$  & std. $0.32\%$ \\
\hline
Weighted Vote & $50.17\%$ & $82.46\%$ & $81.99\%$ & $62.89\%$ & $89.61\%$ &  $63.50\%$  & $70.42\%$  & $81.10\%$ \\
& std. $0.65\%$ & std. $1.54\%$ & std. $0.80\%$ & std. $4.35\%$  & std. $0.83\%$ & std. $2.51\%$  & std. $2.75\%$  & std. $0.72\%$ \\
\hline
Stacking & $14.47\%$ & $41.47\%$ & $70.16\%$ & $66.44\%$  & $89.42\%$ &  $65.06\%$  & $46.27\%$  & $81.95\%$ \\
& std. $1.13\%$ & std. $2.90\%$ & std. $3.35\%$ & std. $3.20\%$ & std. $1.16\%$ & std. $4.97\%$  & std. $3.30\%$  & std. $0.52\%$ \\
\hline
Indep. Copula & $49.19\%$ & $85.77\%$ & $83.21\%$ & $61.38\%$  & $89.70\%$ & $63.89\%$  & $85.45\%$  & $81.56\%$ \\
 & std. $0.64\%$ & std. $1.30\%$ & std. $0.68\%$ & std. $5.82\%$  & std. $1.07\%$ & std. $4.83\%$  & std. $1.31\%$  & std. $2.88\%$ \\
\hline
DELCO & $49.06\%$ & $85.86\%$ & $82.99\%$ & $65.06\%$  & $89.35\%$ & $64.26\%$  & $85.45\%$  & $83.04\%$ \\
Gauss. Copula & std. $0.64\%$ & std. $1.17\%$ & std. $0.83\%$ &  std. $3.01\%$ & std. $1.18\%$ & $4.25\%$  & std. $1.31\%$  & std. $1.68\%$ \\
\hline
Centralized Clf. & $58.19\%$ & $90.65\%$ & $83.16\%$ & $73.83\%$ & $92.26\%$ &  $68.23\%$  & $74.95\%$  & $81.95\%$ \\
& std. $0.36\%$ & std. $0.33\%$ & std. $0.40\%$ &  std. $0.57\%$ & std. $0.52\%$ & std. $0.44\%$  & std. $0.59\%$  & std. $0.52\%$ \\
\hline
\end{tabular}}
\end{table}

In most experiments, decentralized ensemble methods have difficulties to compete with a centralized classifier. This is presumably because PCA-based data splits do not allow to discover better decision frontiers. However, for the Drive and Particle datasets, it is remarkable that the copula-based approaches achieve higher accuracies than the centralized classifier. 


Most importantly, we see that one of the copula-based method is always either the top 1 decentralized method or the top 2 which is in line with the robustness observed in the synthetic data set experiments. When the Gaussian copula is outperformed by the independent copula, the maximal absolute accuracy discrepancy is $0.37\%$. However, when the independent copula is outperformed by the Gaussian one, the maximal absolute accuracy discrepancy is $3.68\%$.

To better upraise the added value brought by DELCO, we performed another experiment in which six out of the ten base classifiers are replaced by six copies of a majority vote ensemble relying on those six base classifiers. In this situation, there is clearly a strong dependency among base classifiers. Since copulas are meant to capture dependency information, a better fit should be achieved by the Gaussian copula. This is indeed confirmed by the corresponding average accuracies which are reported in Table \ref{tab:real4}.

\begin{table}[!t]
\renewcommand{\arraystretch}{1.2}
\caption{Classification accuracies (with standard deviations) for several real data sets. ($m=10$ base classifiers. 6 of them are identical ones.) }
\label{tab:real4}
\centering
\resizebox{\textwidth}{!}{
\begin{tabular}{|c||c|c|c|c|c|c|c|c|}
\hline
 Method & 20newsgroup & MNIST & Satellite & Wine & Spam & Avila & Drive & Particle \\
\hline
Clf. Selection & $45.90\%$ & $73.20\%$ & $79.60\%$ & $62.37\%$ & $86.91\%$ & $58.83\%$  & $64.18\%$  & $81.40\%$ \\
& std. $0.70\%$ & std.  $0.54\%$ & std. $1.03\%$ & std. $4.95\%$ & std. $2.20\%$ & std. $4.88\%$  & std. $2.84\%$ & std. $0.86\%$ \\
\hline
Best Clf.  & $46.11\%$ & $73.26\%$ & $80.23\%$ & $63.30\%$ & $87.42\%$  & $61.24\%$  & $64.31\%$  & $81.68\%$ \\
& std. $0.69\%$ & std. $0.55\%$ & std. $0.75\%$ & std. $4.32\%$ & std. $1.89\%$ & std. $2.98\%$  & std. $2.83\%$  & std. $0.33\%$ \\
\hline
Weighted Vote & $47.07\%$ & $75.14\%$ & $79.79\%$ & $61.87\%$ & $86.94\%$ &  $58.27\%$  & $65.80\%$  & $79.87\%$ \\
& std. $0.66\%$ & std. $0.64\%$ & std. $0.69\%$ & std. $4.62\%$  & std. $2.48\%$ & std. $4.49\%$  & std. $2.63\%$  & std. $1.74\%$ \\
\hline
Stacking & $14.25\%$ & $36.69\%$ & $70.61\%$ & $64.91\%$  & $89.35\%$ &  $61.29\%$  & $40.17\%$  & $80.43\%$ \\
& std. $1.08\%$ & std. $2.13\%$ & std. $3.05\%$ & std. $2.79\%$ & std. $1.47\%$ & std. $4.89\%$  & std. $4.60\%$  & std. $1.43\%$ \\
\hline
Indep. Copula & $47.49\%$ & $76.37\%$ & $81.50\%$ & $62.13\%$  & $87.20\%$ & $58.28\%$  & $71.28\%$  & $81.56\%$ \\
 & std. $0.67\%$ & std. $0.64\%$ & std. $0.87\%$ & std. $4.90\%$  & std. $2.41\%$ & std. $4.50\%$  & std. $2.52\%$  & std. $0.74\%$ \\
\hline
DELCO & $47.04\%$ & $77.97\%$ & $82.00\%$ & $64.65\%$  & $89.43\%$ & $60.93\%$  & $72.10\%$  & $83.15\%$ \\
Gauss. Copula & std. $0.89\%$ & std. $0.62\%$ & std. $0.84\%$ &  std. $2.70\%$ & std. $1.56\%$ & std. $3.48\%$  & std. $2.26\%$  & std. $1.70\%$ \\
\hline
Centralized Clf. & $59.41\%$ & $90.77\%$ & $83.15\%$ & $73.83\%$ & $92.26\%$ &  $61.29\%$  & $74.94\%$  & $88.49\%$ \\
& std. $0.39\%$ & std. $0.14\%$ & std. $3.05\%$ &  std. $0.57\%$ & std. $0.52\%$ & std. $4.88\%$  & std. $0.59\%$  & std. $2.60\%$ \\
\hline
\end{tabular}}
\end{table}

In this second series of results, we see that performance discrepancies between DELCO and the independent copula are much larger. Except for the 20newsgroup data set, the Gaussian copula always achieves higher accuracies than the independent copula. DELCO is the top one decentralized method for 5 datasets and the top 2 for the remaining ones\footnote{We consider that DELCO and weighted vote have equal level of performances for the 20newsgroup data set.}. 
Classifier selection methods are immune to the artificially added dependency because, by construction, they are idempotent methods. They are nevertheless still outperformed by ensemble methods.

\section{Conclusion}
In this paper, we introduce a new ensemble method that relies on a probabilistic model. Given a set of trained classifiers, we evaluate the probabilities of each classifier output given the true class on a validation set. We use a Gaussian copula to retrieve the joint conditional distributions of these latter which allow to build an ensemble decision function that consists in maximizing the probability of the true class given all classifier outputs.

We motivate this new approach by showing that it fits a decentralized learning setting which is a modern concern in a big data context. The approach is validated through numerical experiments on both synthetic and real data sets. We show that a Gaussian copula based ensemble achieves higher robustness than other ensemble techniques and can compete or outperform a centralized learning in some situations.

In future works, we plan to investigate other estimation techniques for the copula parameter than grid search which is suboptimal. In particular, we would like to set up a Bayesian approach to that end. This would also allow us to observe if tying the correlation matrices is too restrictive or not. More complex correlation matrix patterns will also be examined. Also, other copula models will tested and the sensitivity of the method w.r.t the chosen copula family will be studied.
\bibliographystyle{abbrv}
\bibliography{biblio_JK}

\end{document}